\documentclass{article} 
\usepackage[T1]{fontenc}
\usepackage{iclr2027_conference,times}

\usepackage{amsmath,amsfonts,bm}

\def\eqref#1{equation~\ref{#1}}

\def\1{\bm{1}}

\def\va{{\bm{a}}}

\def\vp{{\bm{p}}}

\def\vr{{\bm{r}}}

\def\mF{{\bm{F}}}

\def\mI{{\bm{I}}}

\def\mZ{{\bm{Z}}}

\DeclareMathAlphabet{\mathsfit}{\encodingdefault}{\sfdefault}{m}{sl}
\SetMathAlphabet{\mathsfit}{bold}{\encodingdefault}{\sfdefault}{bx}{n}

\usepackage{hyperref}
\hypersetup{hidelinks}
\usepackage{url}
\usepackage{graphicx}
\usepackage{wrapfig}
\usepackage{booktabs}

\title{Latent RL Priors for Data-Efficient\\Dexterous Manipulation}

\author{%
    \textbf{Yixiao Wang$^{1}$\thanks{This work was done during Yixiao Wang's internship at Amazon.},
    Peiyi Chen$^{2}$, Mustafa Mukadam$^{2}$}\\
    \textbf{Shreekant Gayaka$^{2}$, Masayoshi Tomizuka$^{1}$, Son Tran$^{2}$}\\
    $^{1}$University of California, Berkeley \quad $^{2}$Amazon
}

\newcommand{\method}{Latent RL Prior}

\iclrpreprintcopy 
\begin{document}

\maketitle

\begin{abstract}
Learning dexterous manipulation from limited visual demonstrations is challenging because policies must acquire both perceptual grounding and precise control, while reinforcement learning (RL) in simulation can acquire rich control behaviors from large-scale interaction with privileged information. We introduce \textbf{Latent RL Prior}, a two-stage framework that reuses this control knowledge for data-efficient visual policy learning. During privileged pretraining, we distill large-scale RL rollouts into a flow-matching action decoder conditioned on proprioception and latent task context. During visual adaptation, we freeze the decoder and learn a visual interface that maps images to its latent context using action supervision alone, without privileged labels or latent targets. The same decoder can further be shared across arms by learning coordinated visual contexts for bimanual manipulation. Across three simulated tasks, our method outperforms full-parameter fine-tuning of $\pi_{0.5}$ and GR00T N1.7 in the low-data regime; on door lifting and bimanual assembly, it surpasses full-data baselines using $10\times$ and $5\times$ fewer demonstrations, respectively. Further analysis shows that direct visual-to-latent conditioning improves adaptation performance and training stability over routing visual predictions through the pretrained state encoder. Robustness to unseen mass variation and greater linear accessibility of contact-force information in decoder representations further provide evidence of retained RL-acquired control knowledge.
\ificlrfinal
Project website: \url{https://yixiaowang7.github.io/latent-rl-prior-page/}.
\else
Training and evaluation code and datasets will be released.
\fi
\end{abstract}

\section{Introduction}
\label{sec:introduction}

Dexterous manipulation requires coordinated arm and finger motion under
changing contacts and object dynamics. Learning these behaviors from limited
visual demonstrations is challenging because the policy must acquire both
perceptual grounding and precise action generation
\citep{zhao2023act,chi2023diffusionpolicy}. Pretrained vision-language-action
(VLA) models offer transferable priors from broad visual, language, and robot
datasets \citep{zitkovich2023rt2,kim2024openvla,black2024pi0}. We investigate
a complementary source of prior knowledge for dexterous control: behaviors
acquired through large-scale reinforcement learning (RL) in simulation.

Simulated interaction enables RL policies to acquire manipulation strategies
across diverse object geometries and physical properties
\citep{andrychowicz2020dexterous,singla2024sapg,kedia2026simtoolreal}.
However, these policies often rely on privileged information, such as object
and goal states, that is not directly available at deployment. Reusing their
control capabilities requires an interface that connects visual observations
to the pretrained controller. Our premise is that preserving pretrained
proprioceptive processing and action generation can reduce the visual
demonstrations needed to adapt these capabilities to downstream tasks.

We introduce \method{}, a simple yet effective framework for this transfer.
We distill rollouts from an RL teacher trained on procedural pose alignment
into a flow-matching action decoder \citep{lipman2023flowmatching}. The decoder
receives proprioceptive history through a dedicated pathway and latent context
encoding privileged task information. During visual adaptation, we freeze the
decoder and learn a visual interface that maps images directly to this context.
This \emph{latent transfer} uses only the action objective on
image--proprioception--action demonstrations, without requiring
corresponding privileged information, pretrained latent targets, or paired
trajectories between the pretraining and adaptation datasets.

The same design extends a single-arm RL prior to multiple arms. We share the
frozen decoder across arms and learn coordinated visual contexts, each paired
with the corresponding arm's proprioceptive history. Bimanual adaptation thus
reuses the pretrained control prior while learning inter-arm coordination
through the visual interface.

We evaluate the shared prior on target pose alignment, door lifting, and
bimanual assembly in simulation. With limited visual demonstrations, our
method outperforms full-parameter fine-tuning of $\pi_{0.5}$
\citep{physicalintelligence2025pi05} and GR00T N1.7
\citep{nvidia2026gr00tn17}. On door lifting and bimanual assembly, it surpasses
both baselines trained on the full task datasets with $10\times$ and
$5\times$ fewer demonstrations, respectively. Comparisons with the same visual
policy architecture trained from scratch further support the value of the
pretrained action prior.

Further investigations examine how to access and adapt this prior. Latent
transfer improves performance and training stability over explicit transfer
through the pretrained state encoder. Separate adaptation studies favor
conservative updates to the image encoder and a frozen action decoder,
particularly for precise alignment. We also investigate the control behaviors
and physical information accessible after transfer. The adapted policies
exhibit robustness to mass variation absent from the visual demonstrations,
and their decoder representations make contact-force information more linearly
accessible than those learned from scratch. These observations are consistent
with retained control knowledge from large-scale RL pretraining.

Our contributions are:
\begin{itemize}
    \item a framework for adapting a frozen RL action prior through latent
    visual conditioning with action supervision alone, including reuse of
    a single-arm decoder for multi-arm manipulation;
    \item empirical evidence of improved visual-demonstration efficiency
    across three dexterous tasks, together with evaluations of the transfer
    interface and adaptation choices; and
    \item behavioral and representational analyses of mass robustness and
    contact-force accessibility after visual adaptation.
\end{itemize}

\section{Related Work}
\label{sec:related_work}

\paragraph{Dexterous control from reinforcement learning.}
RL with dynamics randomization has enabled dexterous in-hand manipulation
\citep{andrychowicz2020dexterous}. Parallel simulation and scalable policy
optimization support learning from large amounts of interaction
\citep{makoviychuk2021isaacgym,singla2024sapg}. SimToolReal trains
a single policy to manipulate procedurally generated objects toward goal poses
and deploys it using object-pose estimates and goal-pose sequences
\citep{kedia2026simtoolreal}. We build on its released teacher, distill
rollouts into an action decoder, and investigate reuse of the resulting control
prior through visual conditioning learned from limited downstream
demonstrations.

\paragraph{Transfer with privileged information.}
Asymmetric actor--critic training uses privileged information in the critic
to learn policies with restricted observations \citep{pinto2018asymmetric}.
Policy distillation transfers behavior between policies
\citep{rusu2016policydistillation}, while Learning by Cheating uses a
privileged teacher to train a visual student \citep{chen2020learningbycheating}.
Closer to our interface, Rapid Motor Adaptation (RMA) and its extension to
in-hand rotation retain an RL controller while estimating latent representations
of physical properties from proprioceptive and action histories
\citep{kumar2021rma,qi2023hora}. Their adaptation modules are trained to match
embeddings computed from corresponding privileged information. Our visual
adaptation uses action supervision alone: the decoder distilled from RL
rollouts remains frozen, while its visual conditioning is learned without
privileged information or latent targets in the downstream demonstrations.
We reuse this prior across tasks and extend it to bimanual manipulation
through a shared single-arm decoder.

\paragraph{Visual imitation and generalist robot policies.}
Action chunking and generative action modeling support learning visual
manipulation from demonstrations \citep{zhao2023act,chi2023diffusionpolicy}.
Generalist policies and VLA models transfer knowledge from diverse robot and
vision--language datasets
\citep{zitkovich2023rt2,ghosh2024octo,kim2024openvla}. The $\pi_0$ and GR00T
families combine pretrained vision--language representations with
flow-matching action models \citep{black2024pi0,nvidia2025gr00t}.
Our approach combines DINOv2 visual features \citep{oquab2024dinov2} with a
control prior distilled from privileged RL rollouts. We investigate how this
source of control experience supports adaptation to dexterous tasks with
limited visual demonstrations, comparing against full-parameter fine-tuning
of $\pi_{0.5}$ \citep{physicalintelligence2025pi05} and GR00T N1.7
\citep{nvidia2026gr00tn17}.

\section{Latent RL Prior}
\label{sec:method}

\begin{figure}[t]
    \centering
    \includegraphics[width=\linewidth]{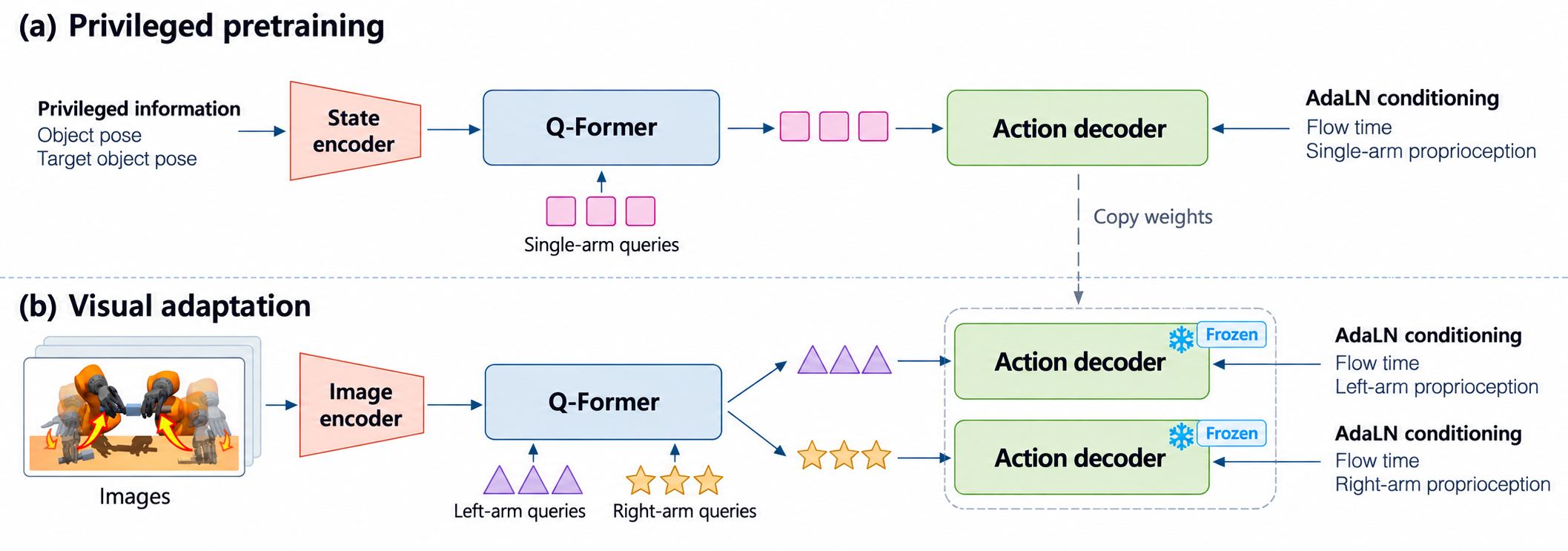}
    \caption{\method{}. (a) Privileged pretraining: a state tokenizer and
    Q-Former supply task context to an action decoder conditioned on flow time
    and single-arm proprioception. (b) Visual adaptation: an image encoder and
    Q-Former learn context tokens while the decoder remains frozen. For
    bimanual assembly, left- and right-arm queries interact within the same
    Q-Former; each output group conditions the shared decoder with its
    corresponding arm's proprioception.}
    \label{fig:method_overview}
\end{figure}

Large-scale reinforcement learning (RL) in simulation can acquire complex dexterous manipulation behavior through abundant interaction across diverse object geometries, masses, and physical properties. However, these policies often rely on privileged state information, such as object and goal states, that is unavailable at deployment. Therefore, in this section, we investigate how to transfer such dexterous behaviors acquired by privileged RL into a reusable \emph{RL prior} for downstream visual tasks. Our goal is to preserve dexterous capabilities learned from large-scale simulated experience while adapting a visual interface from limited demonstrations using only deployable observations.

\subsection{Problem Formulation}
\label{sec:problem}

We consider dexterous manipulation tasks in which, at time $t$, the robot
observes its proprioceptive state $\vr_t$ and synchronized images
$\mI_t^{1:C}$ from $C$ cameras, and executes an action $\va_t$.
Using a history of $H$ observations, the policy predicts a $K$-step action
chunk. We denote
\[
\mathcal{P}_t=\vp_{t-H+1:t},\quad
\mathcal{R}_t=\vr_{t-H+1:t},\quad
\mathcal{I}_t=\mI_{t-H+1:t}^{1:C},\quad
\mathcal{A}_t=[\va_t,\ldots,\va_{t+K-1}].
\]

We represent task information with $M$ latent context tokens
$\mZ_t \in \mathbb{R}^{M \times d}$, each of dimension $d$.
Together with the proprioceptive history $\mathcal{R}_t$, these tokens
condition the action decoder to predict the action chunk $\mathcal{A}_t$.
The source of $\mZ_t$ depends on the training stage: it is derived from
the privileged history $\mathcal{P}_t$ during pretraining and from the
visual history $\mathcal{I}_t$ during visual adaptation.

We use a large privileged RL dataset
$\mathcal{D}_{\mathrm{RL}}=\{(\mathcal{P}_t,\mathcal{R}_t,\mathcal{A}_t)\}$
and a smaller downstream visual dataset
$\mathcal{D}_{\mathrm{vis}}=\{(\mathcal{I}_t,\mathcal{R}_t,\mathcal{A}_t)\}$.
The visual demonstrations contain no privileged-state labels and need not
be paired with the RL trajectories. Our goal is to reuse the action prior
learned from $\mathcal{D}_{\mathrm{RL}}$ while replacing privileged
conditioning with visual conditioning for downstream tasks.

\subsection{Privileged Pretraining}
\label{sec:privileged_pretraining}

\paragraph{Large-scale teacher rollouts.}
Following \citet{kedia2026simtoolreal}, we first train an RL teacher on a broad object pose-alignment task, where the robot aligns procedural objects with
target poses. We apply domain randomization over object shape, size, and
density during training to expose the policy to diverse manipulation
conditions. We then roll out the trained policy at scale and construct
$\mathcal{D}_{\mathrm{RL}}$ from the collected trajectories by extracting
privileged histories $\mathcal{P}_t$, proprioceptive histories
$\mathcal{R}_t$, and future action chunks $\mathcal{A}_t$.

\paragraph{Flow-matching action prior.}
A state tokenizer $T_{\eta}$ followed by a Q-Former $Q_{\psi}$
\citep{li2023blip2} maps the privileged history to the latent context
\[
\mZ_t^{\mathrm{priv}}
=
Q_{\psi}\!\left(T_{\eta}(\mathcal{P}_t)\right)
\in \mathbb{R}^{M\times d}.
\]
The action decoder $v_{\theta}$ is a transformer
\citep{vaswani2017attention} whose action tokens cross-attend to this
context. The proprioceptive history $\mathcal{R}_t$ and flow time $\tau$
condition each block through adaptive layer normalization (AdaLN)
\citep{peebles2023dit}. This separation allows the source of the latent
context to change while preserving the pretrained processing of
proprioception.

We train the decoder with conditional flow matching
\citep{lipman2023flowmatching}. For a demonstrated action chunk
$\mathcal{A}_t^0 := \mathcal{A}_t$, we sample
$\mathcal{A}_t^1 \sim \mathcal{N}(\bm{0},\mathbf{I})$ and
$\tau \sim \mathcal{U}[0,1]$, and form
$\mathcal{A}_t^{\tau}
=(1-\tau)\mathcal{A}_t^0+\tau\mathcal{A}_t^1$.
The decoder learns the corresponding velocity field by minimizing
\begin{equation}
\mathcal{L}_{\mathrm{FM}}
=
\mathbb{E}_{\substack{
(\mathcal{R}_t,\mathcal{A}_t^0)\sim\mathcal{D}_{\mathrm{RL}},\,
\mathcal{A}_t^1\sim\mathcal{N}(\bm{0},\mathbf{I}),\,
\tau\sim\mathcal{U}[0,1]
}}
\left[
\left\|
v_{\theta}\!\left(
\mathcal{A}_t^\tau,\tau,\mZ_t,\mathcal{R}_t
\right)
-
\left(\mathcal{A}_t^1-\mathcal{A}_t^0\right)
\right\|_2^2
\right].
\label{eq:flow_matching}
\end{equation}

\subsection{Visual Adaptation}
\label{sec:latent_adaptation}

Having pretrained the action prior from privileged RL rollouts, we next
adapt it to downstream visual observations, where privileged task information
is unavailable. We learn a visual conditioning module from limited
demonstrations while keeping the pretrained action decoder fixed.
Specifically, an image encoder
$E_{\mathrm{vis}}$, instantiated with DINOv2 \citep{oquab2024dinov2},
encodes the visual history as
$\mF_t = E_{\mathrm{vis}}(\mathcal{I}_t)$.
We keep the pretrained action decoder $v_{\theta}$ fixed and train the visual
pathway on $\mathcal{D}_{\mathrm{vis}}$ using the same flow-matching
objective in Equation~\ref{eq:flow_matching}. The action loss provides the
only supervision; no privileged-state labels or pretrained latent targets
are required.

\paragraph{Latent and explicit transfer.}
We consider two interfaces for conditioning the pretrained action prior from
visual observations. In \emph{latent transfer}, a visual Q-Former
$Q_{\phi}$ maps image features directly to the decoder context,
\[
\mZ_t^{\mathrm{latent}}
=
Q_{\phi}(\mF_t)
\in \mathbb{R}^{M\times d}.
\]
This replaces the pretrained state tokenizer and state Q-Former, while
preserving the action decoder and its proprioceptive conditioning pathway.
We optimize $E_{\mathrm{vis}}$ and $Q_{\phi}$ with
$\mZ_t=\mZ_t^{\mathrm{latent}}$.

In \emph{explicit transfer}, a bridge $B_{\rho}$ first maps visual features
to a privileged-shaped history,
$\widehat{\mathcal{P}}_t = B_{\rho}(\mF_t)$, which is then processed by the
pretrained state encoder. The two interfaces are therefore
\begin{equation}
\mZ_t^{\mathrm{latent}}
=
Q_{\phi}(\mF_t),
\qquad
\mZ_t^{\mathrm{explicit}}
=
Q_{\psi}\!\left(
T_{\eta}(\widehat{\mathcal{P}}_t)
\right).
\label{eq:transfer_interfaces}
\end{equation}
For explicit transfer, we freeze $T_{\eta}$, $Q_{\psi}$, and $v_{\theta}$,
and optimize only $E_{\mathrm{vis}}$ and $B_{\rho}$. 

\paragraph{Extension to multi-arm manipulation.}
Many bimanual manipulation tasks build on dexterous skills that are already
required at the single-arm level, which motivates
reusing the pretrained single-arm \emph{RL prior} for both arms, while
learning the additional inter-arm coordination from bimanual
demonstrations. Rather than introducing a separate bimanual action decoder,
we retain the shared single-arm decoder and extend its visual conditioning
to produce coordinated, arm-specific contexts.

Specifically, as shown in Figure~\ref{fig:method_overview}, we assign $M$ queries to each arm and process the two query
groups jointly with the visual Q-Former. Self-attention between the groups
allows each arm's context to incorporate information from the other arm
before action generation. The resulting contexts
$\mZ_t^{L}$ and $\mZ_t^{R}$ are paired with the corresponding
proprioceptive histories $\mathcal{R}_t^{L}$ and $\mathcal{R}_t^{R}$.
The same frozen decoder $v_{\theta}$ processes the two context--proprioception
pairs separately to generate arm--hand actions for the left and right arms,
respectively. Thus, both arms reuse the action-generation and proprioceptive
processing learned during privileged pretraining, while cross-arm
coordination is introduced through the jointly inferred visual contexts.

\section{Experiments}
\label{sec:experiments}

We first evaluate whether the pretrained action prior improves
visual-demonstration efficiency across downstream dexterous manipulation
tasks. We then compare latent transfer, which maps visual features directly
to decoder context tokens, with explicit transfer, which conditions the
decoder through the pretrained state encoder, in terms of adaptation
performance and training stability. Next, we examine the effects of
fine-tuning the image encoder $E_{\mathrm{vis}}$ and the action decoder
$v_{\theta}$. Finally, we investigate what control behaviors and physical
information are retained from large-scale RL training after visual
adaptation, using mass-variation tests and contact-force probes.

\subsection{Experimental Setup}
\label{sec:experimental_setup}

\begin{figure}[t]
  \centering
  \includegraphics[width=\textwidth]{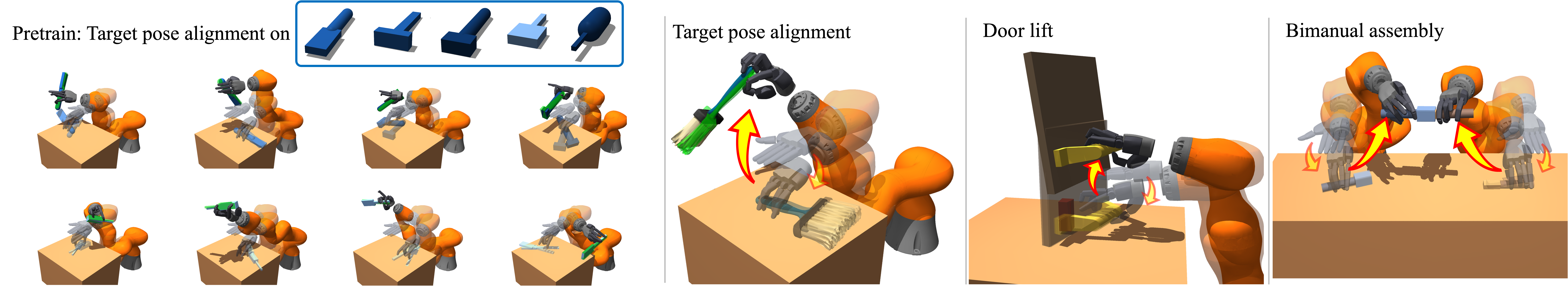}
  \caption{Pretraining and downstream tasks. Left: the pretraining task of
  target pose alignment with varied procedural objects; the inset shows
  example object templates. Right: the downstream tasks of target pose
  alignment with a held-out brush, door lifting, and bimanual assembly
  (left to right). The same pretrained action decoder is reused for visual
  adaptation across all three downstream tasks.}
  \vspace{-12pt}
  \label{fig:task_visualizations}
\end{figure}

\paragraph{Task design and evaluation.}
We conduct the experiments in MuJoCo Warp (MJWarp) \citep{mujocowarp} using
KUKA iiwa arms equipped with SHARPA dexterous hands.
Figure~\ref{fig:task_visualizations} shows the pretraining task
(Section~\ref{sec:privileged_pretraining}) and three downstream tasks.

\begin{figure}[t]
    \centering
    \begin{minipage}{0.33\textwidth}
        \centering
        \includegraphics[width=0.82\linewidth]{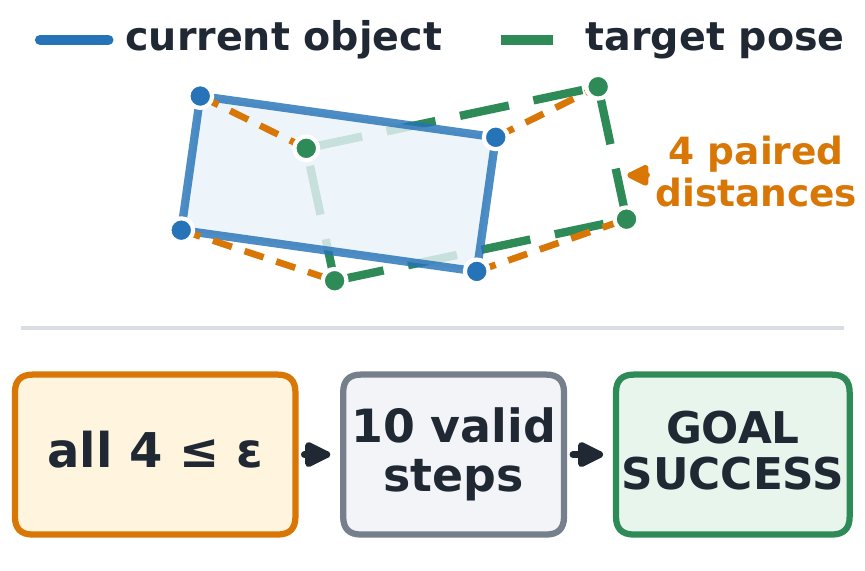}
        \caption{Target pose alignment success criterion.}
        \label{fig:success_definition}
    \end{minipage}\hfill
    \begin{minipage}{0.65\textwidth}
        \centering
        \makeatletter
        \def\@captype{table}
        \makeatother
        \caption{Datasets for privileged pretraining and visual adaptation. F: front; S: side; H: in-hand
        (L/R: left/right).}
        \label{tab:dataset_overview}
        \footnotesize
        \setlength{\tabcolsep}{2.2pt}
        \renewcommand{\arraystretch}{1.15}
        \resizebox{\linewidth}{!}{%
            \begin{tabular}{@{}lrclc@{}}
                \toprule
                \textbf{Dataset} & \textbf{Episodes} &
                \shortstack{ \textbf{Arm--hand} \\ \textbf{systems}} &
                \textbf{Images} &
                \shortstack{\textbf{Privileged}\\\textbf{information}} \\
                \midrule
                Pretraining & 100{,}000 & single & -- & Yes \\
                \midrule
                \shortstack[l]{Target pose alignment} & 4{,}000 & single & F, S, H & No \\
                Door lift & 500 & single & F, S, H & No \\
                \shortstack[l]{Bimanual assembly} & 5{,}000 & bimanual & F, H$_L$, H$_R$ & No \\
                \bottomrule
            \end{tabular}%
        }
    \end{minipage}
\end{figure}

\emph{Pretraining.} Following the SimToolReal experimental setup
\citep{kedia2026simtoolreal}, a single arm--hand system reorients procedural
object templates with randomized shapes, sizes, and densities to target poses.
We measure translation and orientation error using four corresponding
keypoints from the clean object and target poses:
$d_{\max}=\max_{j\in\{1,\ldots,4\}}
\|\mathbf{k}_j^{\mathrm{obj}}-\mathbf{k}_j^{\mathrm{goal}}\|_2$.
Success requires $d_{\max}\leq\epsilon=0.015$~m on 10 policy steps,
not necessarily consecutive (Figure~\ref{fig:success_definition}).

\emph{Downstream tasks.} (1) In \emph{target pose alignment}, the robot grasps
a blue brush excluded from pretraining and aligns it with a target pose. We use
the same keypoint criterion and report strict ($\epsilon=0.015$ m) and relaxed
($\epsilon=0.05$ m) success rates. (2) In \emph{door lifting}, the robot lifts
a door to reveal a cube behind it. We randomize the cube's height, color, and
size, as well as the door's position and orientation. Success requires
lifting the door above the cube's height. (3) In \emph{bimanual assembly}, two
arm--hand systems grasp separate objects from a tabletop, reorient them in
hand, and insert a cubic protrusion on one object into a matching cavity in
the other. We randomize the objects' masses and sizes, as well as the cavity
size. Success requires full insertion, measured by insertion depth relative to
cavity depth. Blue-brush alignment evaluates transfer to an unseen object
within the pretraining task, whereas door lifting and bimanual assembly
evaluate transfer to different tasks.

\paragraph{Data and implementation.}
We fine-tune the released SimToolReal teacher checkpoint
\citep{kedia2026simtoolreal} on procedural target pose alignment and collect
100{,}000 successful rollouts for $\mathcal{D}_{\mathrm{RL}}$. We train a
separate SAPG policy \citep{singla2024sapg} for each downstream task and
collect its a small amount rollouts for $\mathcal{D}_{\mathrm{vis}}$. Table~\ref{tab:dataset_overview} shows the dataset details. Unless otherwise noted, we use latent transfer
(Section~\ref{sec:latent_adaptation}) with DINOv2-Small (ViT-S/14) as
$E_{\mathrm{vis}}$ and reuse the same pretrained action decoder across
downstream tasks.

\paragraph{Baselines.}
We compare our method with (1) full-parameter fine-tuning of the vision-language-action
models $\pi_{0.5}$ \citep{physicalintelligence2025pi05} and GR00T N1.7
\citep{nvidia2026gr00tn17}; (2) a scratch baseline that uses the same visual
policy architecture without the pretrained action prior; and (3) explicit
transfer, as defined in Section~\ref{sec:latent_adaptation}.

\paragraph{Training and evaluation protocol.}
For the main comparisons, we fine-tune GR00T N1.7 and $\pi_{0.5}$ for
30{,}000 optimization steps using their officially recommended fine-tuning
hyperparameters. We train our policy for 100{,}000 steps. For all methods,
we save checkpoints every 5{,}000 steps and retain the final five for
evaluation. We evaluate these checkpoints on held-out environment configurations, which means environment seeds distinct
from those used for dataset generation. Each checkpoint is evaluated
over 256 policy rollouts. Following prior work \citep{wang2024sparse, chi2023diffusionpolicy}, we report
the highest success rate among the five evaluated checkpoints for each run.
Unless otherwise noted, we use three training seeds and report the mean and standard deviation of
these per-run maximum success rates. The training seeds control
demonstration subsampling and the PyTorch random-number generator such as the initialization of newly introduced parameters.

\subsection{Latent RL Prior Enables Data-Efficient Visual Adaptation}
\label{sec:main_results}

\begin{figure}[t]
    \centering
    \includegraphics[width=\linewidth]{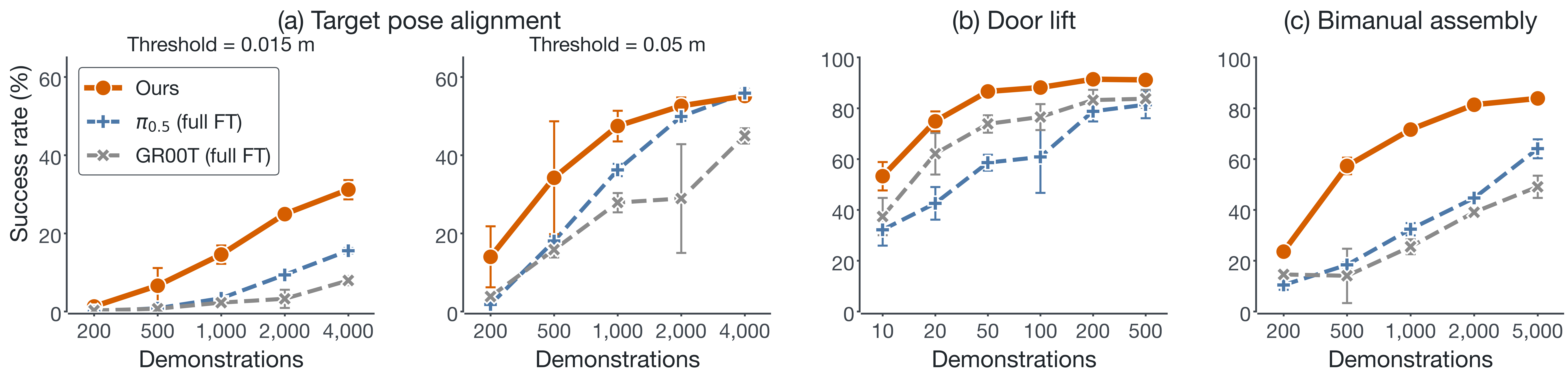}
\caption{Comparison with full-parameter fine-tuning of GR00T N1.7 and
$\pi_{0.5}$ across demonstration budgets. (a) Target pose alignment at strict
($\epsilon=0.015$ m) and relaxed ($\epsilon=0.05$ m) thresholds; (b) door
lifting; (c) bimanual assembly. We report average success rate and standard
deviation across three random seeds.}
\label{fig:data_efficiency_thresholds}
\end{figure}

\paragraph{Comparison with VLA baselines.}
As shown in Figure~\ref{fig:data_efficiency_thresholds}, our method achieves higher success rates than GR00T N1.7 and $\pi_{0.5}$
with limited visual demonstrations across all three tasks.
The advantage is particularly pronounced in the low-data regime.
On door lifting, our method trained with only 50 demonstrations outperforms
both VLA baselines trained on the full 500-demonstration dataset, using
$10\times$ fewer demonstrations. On bimanual assembly, it likewise
surpasses both full-data baselines with 1{,}000, a $5\times$ reduction. For target pose alignment, under the strict $0.015$-m criterion,
our method maintains an advantage across demonstration budgets, consistent
with the pretrained RL prior retaining control capabilities useful for
precise alignment. Under the relaxed $0.05$-m criterion, the advantage
remains clear in the low-data regime but narrows as more demonstrations
become available, with performance becoming numerically comparable to
$\pi_{0.5}$ at 4{,}000 demonstrations. Overall, these results indicate that
the RL prior is most beneficial when visual demonstrations are scarce or
when the task demands precise control.

\begin{wrapfigure}{R}{0.65\textwidth}
    \vspace{-0.8em}
    \centering
    \includegraphics[width=\linewidth]{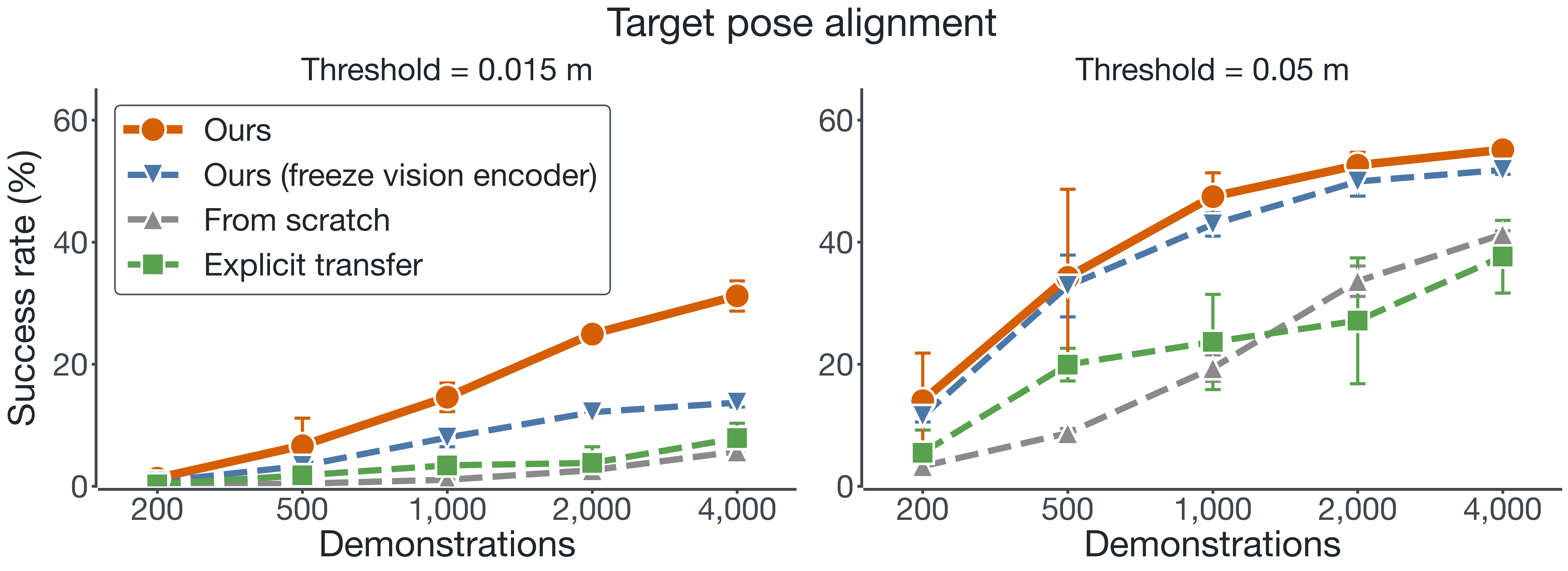}
    \caption{Target pose-alignment ablations with the action decoder frozen for both
latent-transfer variants. Results show mean success rates and standard
deviations over three random seeds, except for explicit transfer, which
uses five seeds. }
    \label{fig:transfer_visual_adaptation}
    \vspace{-0.7em}
\end{wrapfigure}

\paragraph{Contribution of the pretrained RL prior.}
To isolate the contribution of the pretrained RL prior, we compare latent
transfer with training from scratch using the same visual policy architecture
and training setup. As shown in
Figure~\ref{fig:transfer_visual_adaptation}, latent transfer with the image
encoder frozen consistently achieves higher success rates than the
from-scratch baseline across the evaluated demonstration budgets and success
thresholds. Since the visual encoder remains fixed, these gains cannot be
explained by improved visual representation learning during adaptation.
The results therefore support that reusing the pretrained RL action prior
substantially improves learning from limited visual demonstrations, providing
a key source of the data efficiency observed in our method.

\subsection{Latent Transfer Is More Effective under Action-Only Supervision}
\label{sec:transfer_interface}

We compare latent and explicit transfer under action-only supervision,
considering both adaptation performance and training stability.

\paragraph{Adaptation performance.}
As shown in Figure~\ref{fig:transfer_visual_adaptation}, latent transfer
achieves substantially higher target pose-alignment success than explicit
transfer across the evaluated demonstration budgets under both the strict and
relaxed success criteria (note that the image encoder is frozen). These
results favor latent transfer for visual adaptation under action-only
supervision. Interestingly, under the strict $0.015$-m criterion, explicit transfer still
outperforms training from scratch across most demonstration budgets,
suggesting that the pretrained RL prior retains useful capabilities for
precise manipulation even when accessed through the explicit interface.
Under the relaxed $0.05$-m criterion, however, explicit transfer performs
similarly to the from-scratch baseline for much of the data range, with a
clearer advantage only in the low-data regime. This pattern is consistent
with the results in Section~\ref{sec:main_results}, where the benefit of the
RL prior is most pronounced with limited demonstrations and under stricter
precision requirements. More importantly, the substantially stronger and
more consistent performance of latent transfer indicates that effectively
reusing the RL prior depends not only on its presence, but also on how visual
observations interface with the pretrained controller.

\begin{figure}[!htbp]
    \centering
    \includegraphics[width=\linewidth]{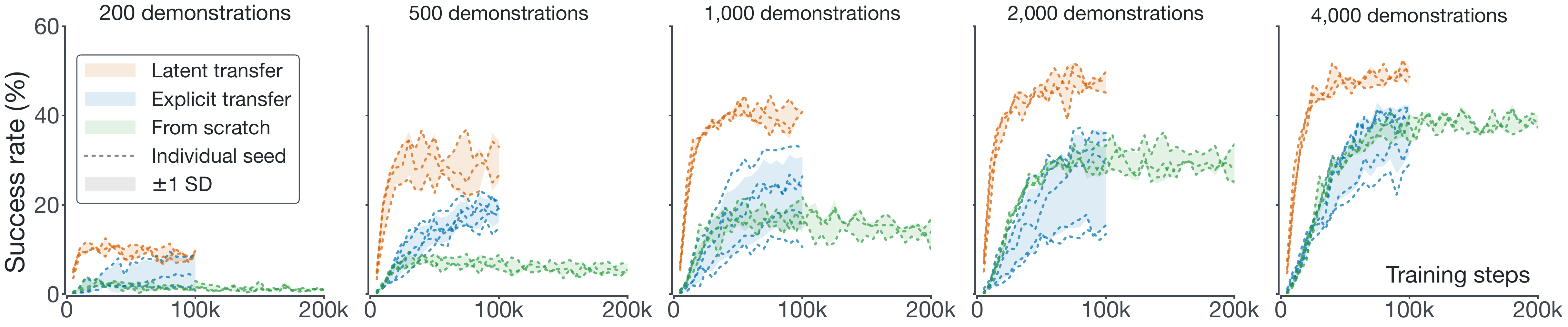}
    \caption{Training stability on target pose alignment at a relaxed ($\epsilon=0.05$ m) threshold. Dotted curves show individual runs for latent transfer, explicit transfer, and scratch training; shaded regions show the mean $\pm$ one standard deviation. Explicit transfer uses five random seeds, while latent transfer and scratch training use three seeds each.}
    \vspace{-12pt}
    \label{fig:method_seed_training}
\end{figure}

\paragraph{Training stability.}
\label{sec:training_stability}
As shown in Figure~\ref{fig:method_seed_training}, with the image encoder
frozen, latent transfer exhibits rapid performance improvement together with
relatively consistent learning trajectories across random seeds. In contrast,
explicit transfer shows greater run-to-run variability, while training from
scratch is comparatively stable but converges to lower performance.
Thus, latent transfer combines consistent learning with stronger performance.

\subsection{Ablations on Visual and Action Adaptation}
\label{sec:finetuning_ablation}

To examine how visual adaptation should modify the pretrained policy, we
independently sweep the learning rates of the image encoder
$E_{\mathrm{vis}}$ and action decoder $v_{\theta}$, while keeping the visual
Q-Former $Q_{\phi}$ trainable throughout.

\begin{figure}[t]
    \centering
    \includegraphics[width=\linewidth]{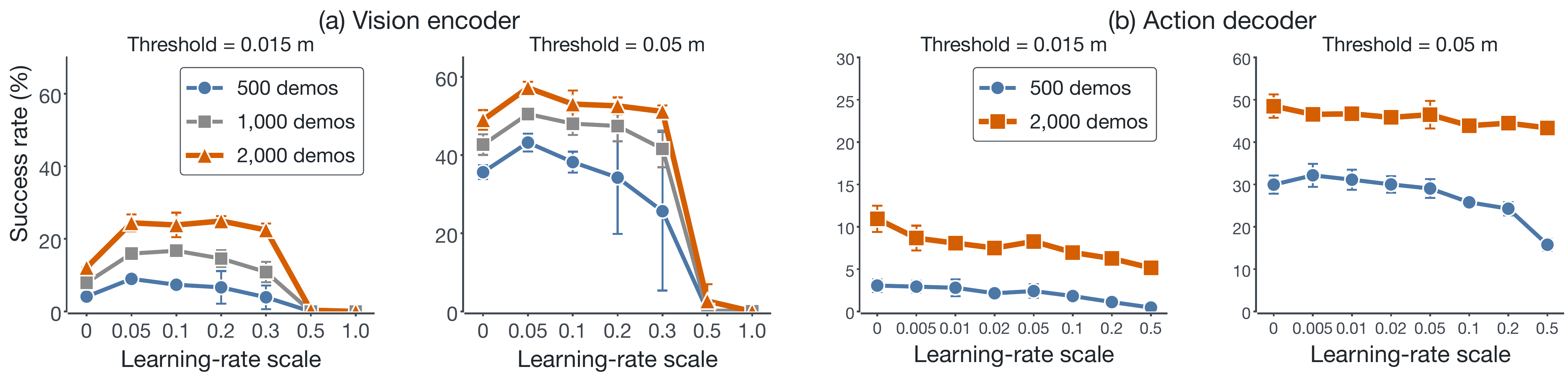}
    \caption{Learning-rate ablations for (a) the image encoder with the action
    decoder frozen and (b) the action decoder with the image encoder frozen.
    Panels report strict ($\epsilon=0.015$ m) and relaxed ($\epsilon=0.05$ m)
    pose-alignment success. We report average success rate and standard deviation across three random seeds. Each
    scale multiplies the tested module's original learning rate; zero freezes it.}
    \label{fig:finetuning_ablation}
\end{figure}

\paragraph{Image-encoder adaptation.}
As shown in Figure~\ref{fig:finetuning_ablation}(a), with the action decoder
frozen, conservative updates to the image encoder improve strict
pose-alignment success across demonstration budgets, whereas larger updates
degrade performance. This effect is particularly pronounced in the low-data
regime: with 500 demonstrations, increasing the learning-rate scale leads to
both lower mean success and greater variation across seeds, while the
1{,}000- and 2{,}000-demonstration settings are comparatively less sensitive.
Under the relaxed criterion, the benefit of image-encoder adaptation is less
uniform. These results suggest that limited-data adaptation benefits from
small, conservative updates to the visual representation, especially when
precise alignment is required.

\paragraph{Action-decoder adaptation.}
As shown in Figure~\ref{fig:finetuning_ablation}(b), fine-tuning the action
decoder provides no benefit across the tested learning-rate scales and
demonstration budgets. Keeping the pretrained decoder frozen performs best
under the strict criterion, while larger updates reduce performance.
We hypothesize that fine-tuning the decoder on a much smaller downstream
dataset may lead to overfitting and reduce the benefit of the control prior
learned from large-scale privileged pretraining with $\mathcal{D}_{\mathrm{RL}}$ in Table~\ref{tab:dataset_overview}. These results favor keeping the
action decoder frozen during visual adaptation.

\subsection{Control Behaviors and Physical Information}
\label{sec:rl_prior_analysis}

Having shown that the pretrained RL prior enables data-efficient visual
adaptation, we next investigate what control behaviors and physical
information from large-scale RL pretraining remain after visual adaptation.
These retained properties may help explain the observed data-efficiency gains.

\begin{wrapfigure}{r}{0.5\textwidth}
    \vspace{-0.8em}
    \centering
    \includegraphics[width=\linewidth]{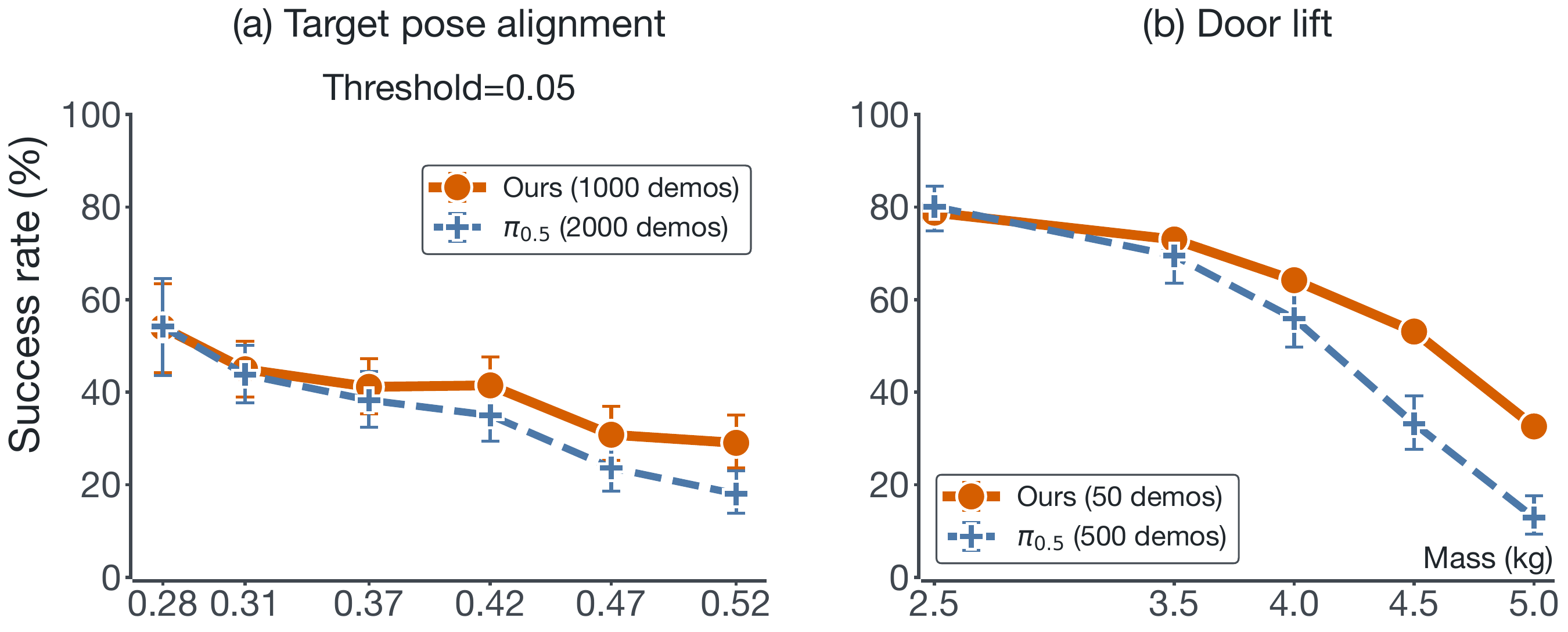}
    \caption{Generalization to increased object mass. (a) Target pose alignment at a relaxed ($\epsilon=0.05$ m) thresholds. (b) Door lift. The leftmost point in
    each plot corresponds to the reference mass used for downstream visual
    training, e.g., for $\mathcal{D}_{\mathrm{vis}}$.}
    \label{fig:mass_robustness}
    \vspace{-0.7em}
\end{wrapfigure}

\paragraph{Generalization to increased object mass.}
RL pretraining includes randomization over object density, whereas the visual
demonstrations contain only a single object mass for each task. We therefore
test whether generalization to mass variation learned during RL pretraining remains
after visual adaptation by evaluating the policies on increasingly heavy
objects without further training
(see results in Figure~\ref{fig:mass_robustness}). For door lifting, our policy uses only
50 demonstrations compared with 500 for $\pi_{0.5}$; for pose alignment, the
corresponding budgets are 1{,}000 and 2{,}000. At the reference mass, the two
methods achieve similar performance. As mass increases, however, our policy
degrades more gradually: for example, on door lifting, it retains 32.6\% success at
5.0~kg compared with 12.9\% for $\pi_{0.5}$. These
results provide evidence that robustness acquired during RL pretraining can
remain useful after visual adaptation.

\begin{figure}[!htb]
    \centering
    \includegraphics[width=\linewidth]{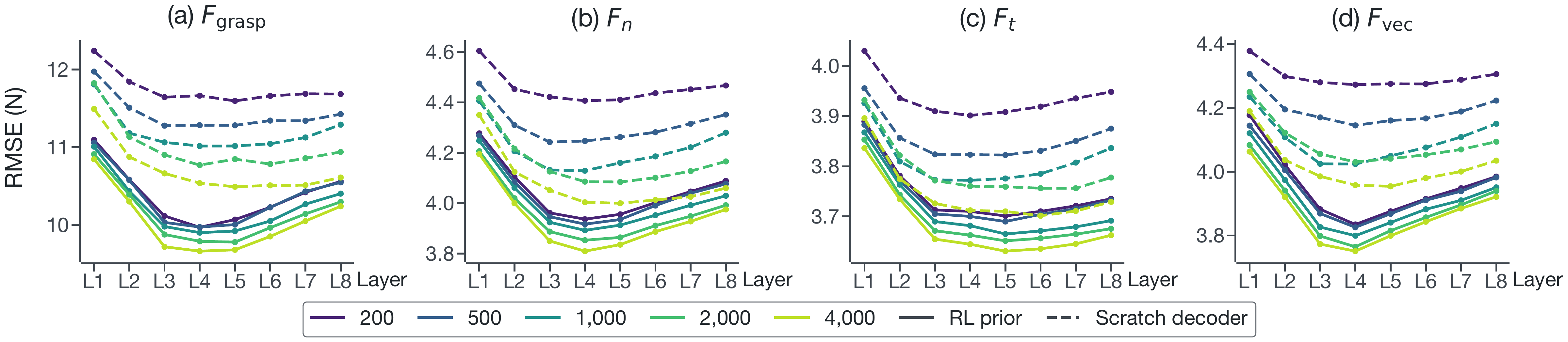}
    \caption{Linear contact-force probes of the visual policies. Panels predict
    (a) total grasp normal force, (b) per-finger normal forces,
    (c) per-finger tangential forces, and (d) per-finger Cartesian force vectors
    from each layer of the action decoder. Colors denote visual-demonstration
    budgets; solid and dashed curves denote the frozen pretrained and
    trainable scratch action decoders. Both policies fine-tune the image
    encoder at $0.2\times$ its base learning rate for 100k steps. Curves
    average three policy seeds on the same IID
    probe test episodes.}
    \label{fig:contact_force_probes}
\end{figure}

\paragraph{Linear probing of contact-force information.}
Contact forces are central to dexterous manipulation, and large-scale
interaction during privileged RL pretraining may induce representations that
capture contact-related physical information. To examine whether such
information is retained after visual adaptation, we probe the hidden
representations of the action decoder for contact forces. Specifically, for
target pose alignment, we collect a separate force-annotated dataset
$\mathcal{D}_{\mathrm{probe}}$ and fit linear ridge probes to the hidden state
at each decoder layer, while keeping the adapted policy frozen. We compare
the RL-prior policy with the same architecture trained from scratch. As shown in Figure~\ref{fig:contact_force_probes}, contact-force information
is substantially more linearly decodable from the RL-prior representations:
the RL-prior model achieves lower probe RMSE than the scratch model across
all evaluated layers and demonstration budgets. More notably, with only 200
visual demonstrations, the RL-prior model already matches or outperforms the
scratch model trained with 4{,}000 demonstrations across the force targets,
with substantial margins on most metrics. This result suggests that
large-scale privileged RL pretraining implicitly learns structured
contact-related physical representations that are retained after visual
adaptation, but are difficult to acquire from limited visual demonstrations
alone. Such retained physical knowledge may therefore contribute to the
strong data efficiency provided by the RL prior.

\section{Limitations}
\label{sec:limitations}

Our results concern downstream visual-demonstration efficiency, not total data
or compute efficiency: the method relies on a capable privileged teacher and
substantial simulated experience. The frozen prior may inherit the teacher's
failure modes and restrict adaptation when new tasks require different control
behaviors. All tasks use the same arm--hand system; bimanual transfer reuses
the single-arm decoder and does not establish transfer to different robot
morphologies. The evaluation is entirely in simulation, including visual
demonstration collection, so physical-robot performance remains untested.
Finally, mass robustness and force readout support the
proposed interpretation but do not isolate a causal mechanism for the
demonstration-efficiency gains.

\section{Conclusion}
\label{sec:conclusion}

We presented \method{}, a simple yet effective framework that transfers RL
control behavior to visual dexterous manipulation by adapting latent
conditioning around a frozen action decoder. This adaptation uses action
supervision alone, without privileged information in the visual demonstrations,
and sharing the decoder extends a single-arm prior to bimanual manipulation.
Across three simulated tasks, the method improves visual-demonstration
efficiency, surpassing the evaluated VLA baselines trained on the full task
datasets with $10\times$ and $5\times$ fewer demonstrations on door lifting and
bimanual assembly, respectively. Further investigations favor latent transfer
over the tested explicit interface for performance and training stability,
while separate adaptation studies favor conservative visual updates and a
frozen decoder. Robustness to mass variation absent from visual demonstrations
and greater linear accessibility of contact-force information than in policies
trained from scratch provide evidence consistent with retained control
knowledge. These findings support preserving RL control priors and adapting
their visual interfaces for data-efficient dexterous manipulation.

\clearpage

\subsection*{AI Use Statement}

We used generative AI tools to help draft and refine manuscript text based on
author-provided drafts and an author-defined narrative, and to assist with
coding and figure preparation. We did not use AI to propose research ideas,
design experiments, or analyze or interpret results. The manuscript was
reviewed by the authors, who take responsibility for the final content.

\subsection*{Ethics Statement}

This work studies dexterous manipulation for robotics. No human data are used.

\subsection*{Reproducibility Statement}

The experiment results are evaluated using at least three random seeds. The main text describes the method and experimental setup for reproducibility. We plan to publicly release the training and evaluation code and datasets.

\bibliography{iclr2027_conference}
\bibliographystyle{iclr2027_conference}

\end{document}